\documentclass[runningheads]{llncs}

\usepackage{booktabs}
\usepackage[table,xcdraw]{xcolor}
\usepackage{graphicx}
\usepackage{subcaption}
\usepackage{cite}
\usepackage{svg}
\usepackage[hidelinks]{hyperref}
\usepackage{comment}
\usepackage{caption}

\makeatletter
\newcommand\footnoteref[1]{\protected@xdef\@thefnmark{\ref{#1}}\@footnotemark}
\makeatother

\begin{document}

\title{Weakly Supervised Attended Object Detection Using Gaze Data as Annotations}
\titlerunning{Weakly Supervised Attended Object Detection Using Gaze Data}
%

\author{Michele Mazzamuto\inst{1}\and
Francesco Ragusa\inst{1,2}\and
Antonino Furnari\inst{1,2}\and \\
Giovanni Signorello\inst{3}\and
Giovanni Maria Farinella\inst{1,2,3}}
\authorrunning{M. Mazzamuto et al.}

%
\institute{FPV@IPLAB, DMI - University of Catania, IT \and
Next Vision s.r.l. - Spinoff of the university of Catania, IT \and 
CUTGANA - University of Catania, IT }

\maketitle              
\begin{abstract}

We consider the problem of detecting and recognizing the objects observed by visitors (i.e., attended objects) in cultural sites from egocentric vision. A standard approach to the problem involves detecting all objects and selecting the one which best overlaps with the gaze of the visitor, measured through a gaze tracker. Since labeling large amounts of data to train a standard object detector is expensive in terms of costs and time, we propose a weakly supervised version of the task which leans only on gaze data and a frame-level label indicating the class of the attended object. To study the problem, we present a new dataset composed of egocentric videos and gaze coordinates of subjects visiting a museum. We hence compare three different baselines for weakly supervised attended object detection on the collected data. Results show that the considered approaches achieve satisfactory performance in a weakly supervised manner, which allows for significant time savings with respect to a fully supervised detector based on Faster R-CNN. To encourage research on the topic, we publicly release the code and the dataset at the following url: \url{https://iplab.dmi.unict.it/WS_OBJ_DET/}

\end{abstract}
\section{Introduction}
Egocentric vision offers a convenient setting to collect visual information from the user’s point of view which can be leveraged to understand human behavior~\cite{Furnari2016474} and intent \cite{Furnari20214021}. The collected information can be in turn used to enable user-centered applications capable of assisting the camera wearer, e.g., providing information about the objects present in the scene or the potential activities which can be initiated. Wearable devices, such as smart glasses, allow to acquire different signals including RGB, depth and gaze, which is the pupil's fixation point of the user in the image.
 Understanding where the user is looking allows to gather insights into their behavior and intention. For example, in the industrial domain, an intelligent system can analyze the attention of the worker to provide information on how to use a specific tool (e.g., for continuous training) or to anticipate which objects they will interact with.
 In the context of healthcare, gaze data can be used to rehabilitate patients with cognitive or attention deficits.
In the cultural heritage domain, wearable devices can be used to improve the fruition of artworks and the visitors experience~\cite{ragusa2019egocentricpoint}. For instance, knowing that the visitor of a museum is looking at a specific artwork gives information on the ongoing activity (observing that artwork), as well as the user’s intention (e.g., the user may want to receive information on the artwork or a specific detail of it). Moreover, the collected information about the attended objects can be useful to better manage cultural sites.

A common approach to detect attended objects in the presence of gaze data consists in running a regular object detector and select the object which best overlaps with gaze data. However, this supervised approach requires a significant quantity of images labeled with bounding boxes around all objects in the scene in order to train the object detector. The involved annotation process is expensive in terms of costs and time. 
\begin{figure}[t]
  \centering
  \includegraphics[width=0.9\linewidth]{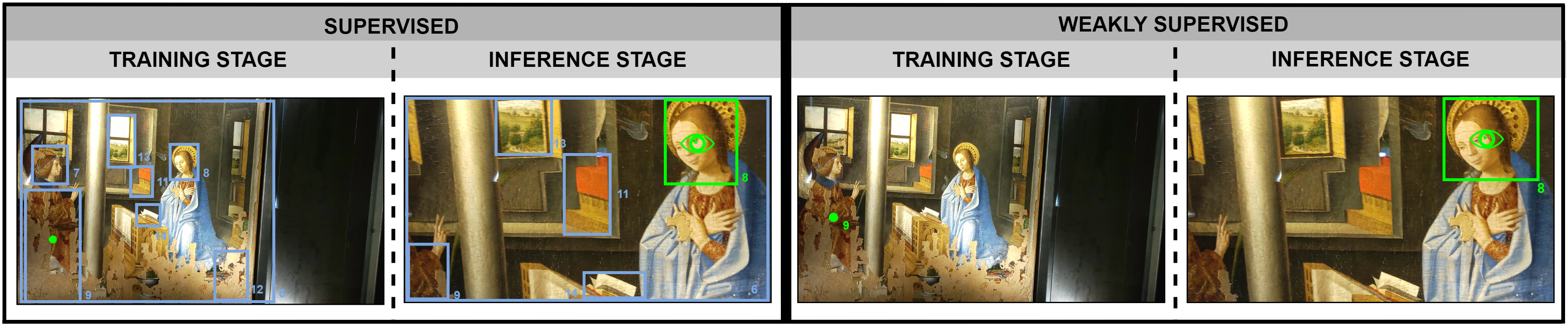}
  \caption{Training and inference stages of both a supervised (left) and a weakly supervised (right) approach. In the supervised approach based on object detection and gaze signal, bounding box annotated around all objects, object classes associated to each box (numbers next to the boxes in the image) and information about the attended object (gaze, represented as a green dot) are both needed. In the weakly supervised setting, only gaze and per-frame labels (numbers next to the green dots) are needed. 
  }
  
  \label{fig:intro_image}
\end{figure}
To streamline data collection and annotation, in this paper we propose to predict attended objects in a weakly supervised way, by relying only on gaze data and a general frame-level label about the currently attended object, which is much cheaper to obtain. Figure \ref{fig:intro_image} summarizes the two approaches.

\begin{figure}[t]
  \centering
  \begin{subfigure}[b]{0.40\linewidth}
    \includegraphics[width=\linewidth]{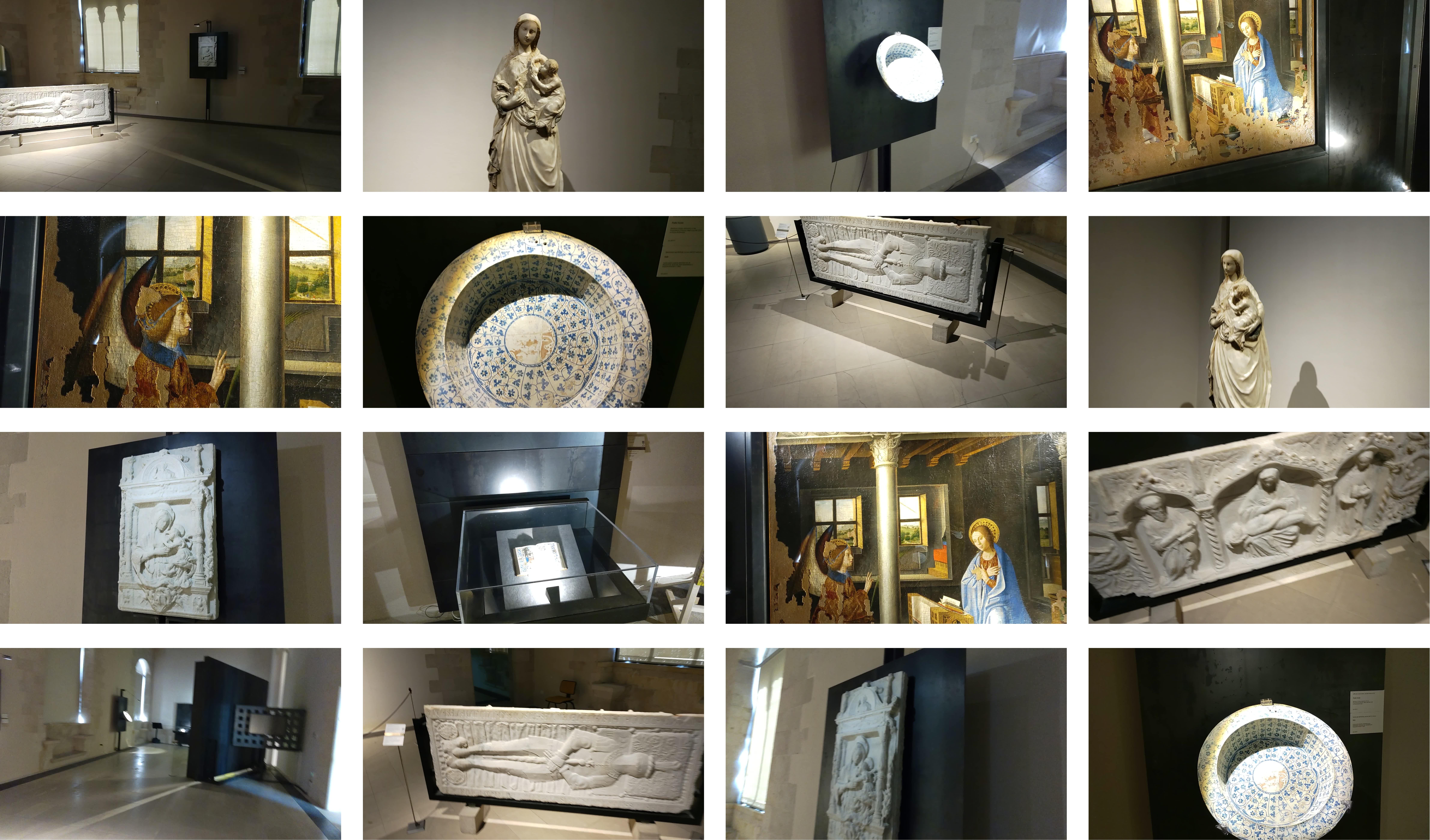}
    \caption{Sample frames}
  \end{subfigure}
  \begin{subfigure}[b]{0.40\linewidth}
    \includegraphics[width=\linewidth]{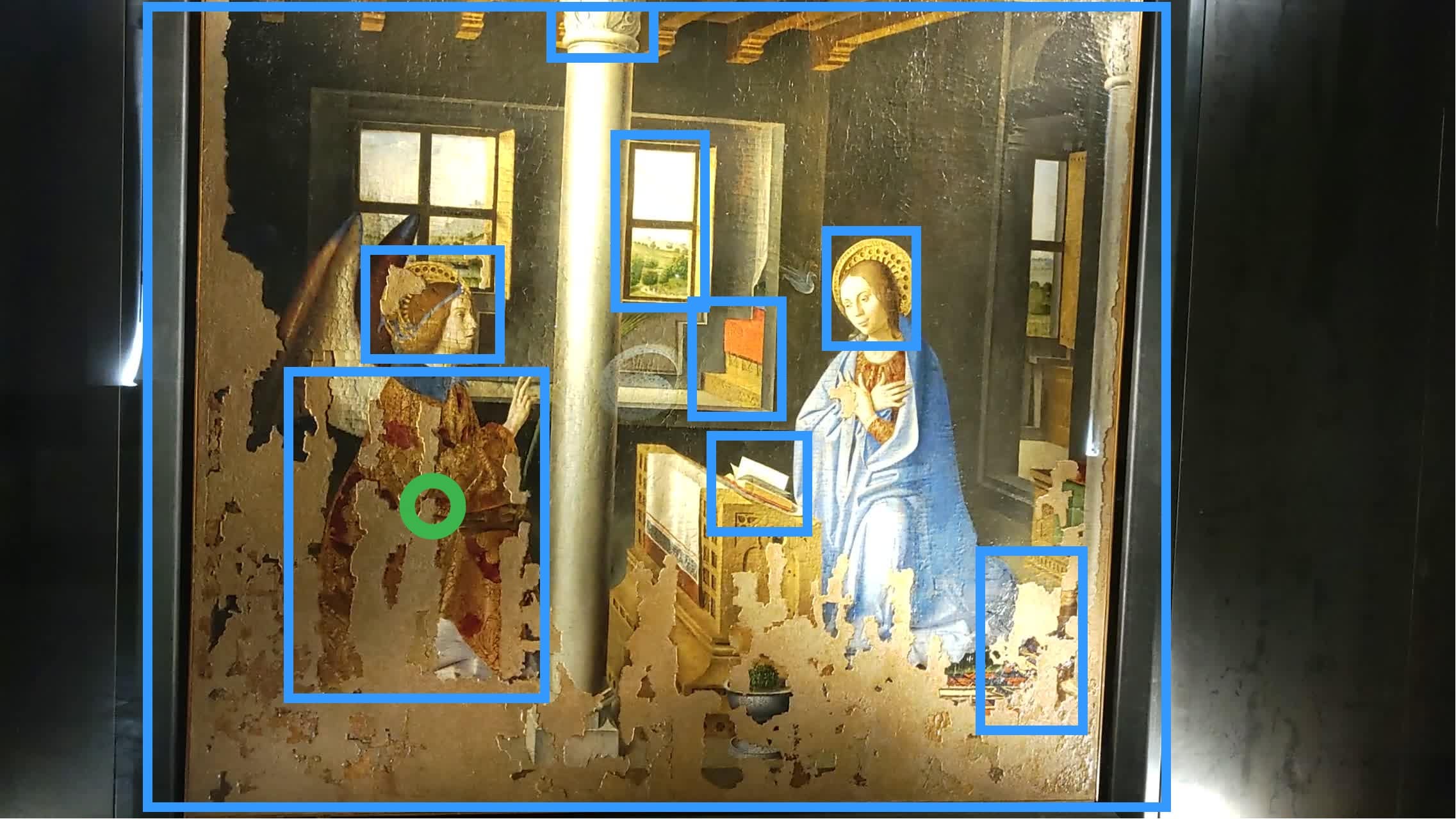}
    \caption{Example annotations}
  \end{subfigure}
  \caption{(a) Sample frames of the proposed dataset  and (b) Examples of bounding box annotations of the considered objects of interest. The green circle represents the gaze of the visitor.}
  \label{fig:dataset}
\end{figure}

To study the problem, we propose a dataset of images collected by visitors in a cultural site, paired with gaze data and frame-level labels about the attended objects. Frames are also labeled with object bounding boxes for comparison with respect to supervised approaches. Figure \ref{fig:dataset}(a) reports some images sampled from the proposed dataset. Apart from the dataset, we investigate and compare three methods which can be trained in a weakly supervised manner using only gaze data and frame-level labels. Experiments show that the fully-convolutional baseline obtains competitive results at a much lower labeling cost. The proposed dataset and the code are publicly available at the following url: \url{https://iplab.dmi.unict.it/WS_OBJ_DET/}. To summarize, the main contributions of this paper are:  1) a new publicly available dataset to study weakly supervised attended object detection and 2) a weakly supervised benchmark for attended object detection.

The remainder of the paper is organized as follows. In Section \ref{related}, we discuss related work. Section \ref{dataset_section} reports the proposed dataset. Section \ref{approach_section} presents the considered approaches, whereas Section \ref{experiments_section} discusses the obtained results. Section \ref{conclusion_section} concludes the paper and
summarises the main accomplishments of our study.

\section{Related Works}
\label{related}

Our work is related to different lines of research: egocentric vision in cultural
sites, weakly supervised object detection and segmentation and analysis of gaze data. The
following sections discuss some relevant works related to these research lines.

\subsection{Datasets in cultural sites}
 There are some public datasets in the cultural heritage context. The OpenMIC~\cite{koniusz2018museum} dataset contains photos captured in ten different exhibition spaces of several museums. The authors of~\cite{koniusz2018museum} used the OpenMIC dataset to explore the problem of artwork identification. NoisyArt~\cite{NoisyArt} is a dataset composed of artwork images collected from Google Images and Flickr containing also metadata (e.g., artwork title, comments, description and creation location) gathered from DBpedia. A recent work presented the AQUA\cite{garcia2020dataset} (Art QUestion Answering) dataset which contains question-answer pairs automatically generated using state-of-the-art question generation methods on the basis of paintings and comments provided by the SemArt~\cite{garcia2020dataset} dataset. EGO-CH~\cite{EGO-CH} is a dataset of egocentric videos for visitors’ behaviour understanding in cultural sites. The dataset has been proposed to study four tasks related to visitors’ behaviour understanding, including point of interest/object recognition task, room-based localization, object retrieval and survey prediction.
 
 Unlike the aforementioned works, we propose a dataset composed of egocentric videos and gaze signal, which is suitable to tackle the tasks of both fully and weakly supervised attended object detection in cultural sites.

\subsection{Object Detection and its use in cultural sites}
Object detection is the task of detecting and recognizing all the objects present in an image. State-of-the-art methods can be grouped into two main groups: two-stage methods and one-stage methods.
Two-stage detectors~\cite{rcnn,fastrcnn,ren2016faster} first predict candidate bounding boxes and then extract visual features from each box to perform the classification of the object class and regress accurate bounding box coordinates. One-stage detectors~\cite{liu2016ssd,retinanet,redmon2018yolov3} predict bounding boxes and classify them in a single step without the need of producing candidate bounding boxes. These detectors usually prioritize the speed of the detection at the expense of prediction accuracy.

Cultural sites generally include both three-dimensional (e.g., statues or artifacts) and two-dimensional (e.g., paintings or book pages) objects, which are often exposed in particular light conditions and protected by glass cases.
The authors of~\cite{10.1145/3092832} designed a smart audio guide that adapts to the actions and interests of museum visitors using a YOLO object detector~\cite{redmon2016look} which localizes and recognizes all objects in the images collected through a smartphone. 
The authors of~\cite{ragusa2019egocentricpoint} investigated the problem of detecting and recognizing points of interest in cultural sites relying on a 2D CNN and a YOLO object detector. The authors of~\cite{PATHAK202174} propose a pipeline to detect damages in images of heritage sites, and subsequently localize the detected damage in the corresponding 3D models. The authors of~\cite{jimaging8020018} present a new dataset called MMSD for medieval musicological studies, which has been manually annotated by a group of musicology experts. The paper also presents a method for black box few-shot object detection.
 
The discussed works are mainly devoted to find all objects in the image following a supervised approach, whereas in this work, we want to detect the object the user is looking at (i.e., the attended object) in a weakly-supervised fashion. All the approaches based on object detection need data labeled with bounding boxes which is domain-specific (i.e., specific to the given cultural site) and hence hard to obtain. To reduce the annotation burden, we explore three different baselines for attended object detection which do not rely on bounding boxes in the training stage.

\subsection{Weakly Supervised Object Detection and Segmentation}
Weakly supervised learning (WSL) approaches are exploited when incomplete or imperfect information is available at training time. These approaches need to operate weakly-labeled training data which allow to reduce the time and cost needed for the data annotation process.
To reduce the annotation time for semantic segmentation and object detection, recent works have focused on training models in a weakly- or semi-supervised setting, such as points~\cite{Bearman2016WhatsTP,2021, cheng2021pointlysupervised} and free-form squiggles\cite{Bearman2016WhatsTP}. In particular, the work of~\cite{cheng2021pointlysupervised} combines small portions of box annotations and points uniformly sampled for each bounding box. 

Recent successes in semantic segmentation have been driven by methods that train CNNs originally designed for image classification to assign semantic labels to each pixel of an image~\cite{long2015fully,6338939,hariharan2014simultaneous,chen2016semantic}.
Inspired by~\cite{long2015fully,zhou2015learning}, we investigate three weakly supervised semantic segmentation approaches to tackle the attended object detection task. 
Importantly, differently from past works~\cite{yoo2019pseudoedgenet,Bearman2016WhatsTP,cheng2021pointlysupervised,Papadopoulos2014TrainingOC}, we explored the attended object detection task without using bounding box annotations~\cite{Papadopoulos2014TrainingOC,2021}. Our experiments show that it is possible to tackle the addressed task using only gaze data as supervision, thus reducing annotation time and costs.

\subsection{Analysis of Gaze Data}
In recent works, gaze data has been increasingly used to tackle various computer vision tasks.
Gaze-tracking data has been used to perform weakly supervised training of object detectors~\cite{6751187,DBLP:conf/cvpr/YunPSZB13}, to infer the semantics of the scene~\cite{See_Better}, to detect salient objects in images~\cite{6909437} and videos~\cite{7298944} or to perform semantic segmentation of images~\cite{5459254}.
Typically, training an object detector requires a large set of images annotated with bounding boxes, which is an expensive and time consuming step. The work of~\cite{Papadopoulos2014TrainingOC} can reduce the annotation time by defining a model that takes human eye movement data as input and infers the spatial support of the target object by labeling each pixel as either object or background.

In this work, we propose to use gaze data and a frame-based label as the only supervision. The gaze signal, which represents the user's attention,  indicates the rough position of the object attended  by the visitor, which we leverage for training.

\section{Dataset}
\label{dataset_section}
\begin{figure}[t]
  \centering
  \begin{subfigure}[b]{0.47\linewidth}
    \includegraphics[width=\linewidth]{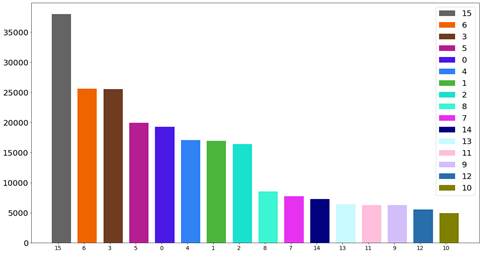}
    \caption{Distribution of attended objects}
  \end{subfigure}
  \begin{subfigure}[b]{0.44\linewidth}
    \includegraphics[width=\linewidth]{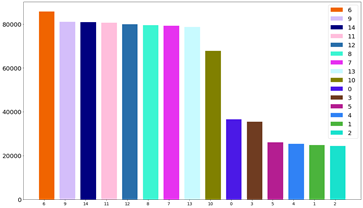}
    \caption{Distribution of all objects}
  \end{subfigure}
  \caption{Dataset statistics.}
  \label{fig:dataset_stats}
\end{figure}

We asked 7 subjects (aged between 24 and 40) to capture egocentric videos while visiting a cultural site containing 15 objects of interest. Videos have been acquired using a head-mounted Microsoft HoloLens2 device\footnote{\url{https://www.microsoft.com/it-it/hololens}} in the room V of the Palazzo Bellomo located in Siracusa. In order to acquire the videos and the gaze coordinates, we developed a HoloLens2 application based on Unity. The developed application allows two acquisition modalities: guided tour and free tour. The guided tour modality guides the user during the visit, generating a path to follow and providing additional audio information for each of the artworks of the site. Instead, the free tour modality leaves the visitor free to explore the environment with no particular structure or restriction about the duration of the visit\footnote{\label{seesup}See supplementary material for more details.}. The egocentric videos have been captured by the subjects who followed the two acquisition modalities. In this way, we acquired 14 videos with a resolution of $2272\times1278$ pixels at 30 fps. The average duration of the videos is 11.25 minutes for the guided tour modality, and 5.66 minutes for the free tour modality. 
We considered 15 objects of interest\footnoteref{seesup}. Note that 8 of the considered objects of interest represent details of the artwork ``Annunciazione".
Figure \ref{fig:dataset}(a) shows some sample frames of the proposed dataset. Each frame of the dataset is associated to 2D coordinates (x,y) relative to the gaze of the visitor, as well as to a class related to the observed (attended) object or a negative class ``other" if the user was not looking at any of the considered objects. The dataset contains 178,977 frames, which we manually labeled with bounding box annotations related to 15 objects of interest for evaluation purposes. Figure \ref{fig:dataset}(b) shows some examples of annotations. The distribution of the considered attended objects is shown in Figure~\ref{fig:dataset_stats}(a), whereas the distribution of all objects appearing in the dataset is shown in Figure~\ref{fig:dataset_stats}(b).

\section{Methods}
\label{approach_section}
 
Differently from the standard object detection task which aims to detect and recognize all the objects present in the scene, attended object detection consists in detecting and recognizing only the object observed by the camera wearer. Formally, let $O = \{o_1, o_2, ..., o_n\}$ be the set of objects in the image and let $C = \{c_1, c_2, ..., c_n\}$ be the corresponding set of object classes. The proposed task consists in detecting the attended object $o_{att}$, predicting its bounding box coordinates $(x_{att},y_{att},w_{att},h_{att})$ and assigning them the correct class label $c_{att}$. We have investigated three different approaches to predict the attended objects in a weakly supervised manner, by relying only on gaze data and a frame level label of the attended object. The approaches are described in the following sections.

\subsection{Sliding Window approach}
\label{Sliding_Window_Approach}
\begin{figure}[t]
  \centering
    \includegraphics[width=0.9\linewidth]{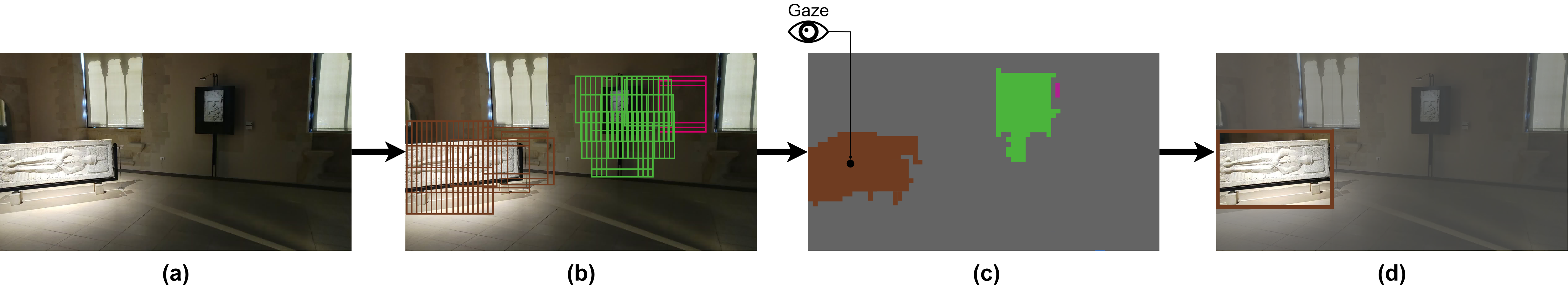}
  \caption{Sliding window approach example.}
  
  \label{fig:slidingnewimage}
\end{figure}

We first investigate a simple approach consisting in exploiting a ResNet18 CNN to obtain a semantic segmentation mask by classifying each image patch to infer whether it belongs to one of the objects of interest or to none of them (``other"). We first train the CNN to classify image patches of size $300\times300$ pixels sampled around gaze coordinates. Each image patch is classified using the frame-level annotations associated to the fixation points (hence the attended object class or a negative “other” class if no object is observed). Note that, in order to optimize this algorithm, only fixation points and temporal annotations indicating the attended objects are needed (e.g. ``the user has looked at object \(c\) from frame \( n\) to frame \(n+m\)'').
At test time we classify all image patches (with a size of $300\times300$ pixels) within the image using a sliding window. The result is a segmentation mask where each element is an integer between 0 and 15 (the ID of the considered classes). Figure~\ref{fig:slidingnewimage}(b) shows an example of classified windows (colors indicate predicted labels) for the input image shown in Figure~\ref{fig:slidingnewimage}(a), whereas Figure~\ref{fig:slidingnewimage}(c) reports the related segmentation mask. While image patches are analyzed at a pre-defined scale, objects may be present at multiple scales, depending on the distance between the observer and the objects. To correctly draw a box around the attended object, we consider a patch of $100\times100$ pixels centered at the 2D gaze coordinates and fit a box to the connected component of the median class of the patch.

\subsection{Fully Convolutional attended object detection}
\label{conv}
\begin{figure}[t]
  \centering
  \includegraphics[width=0.8\linewidth]{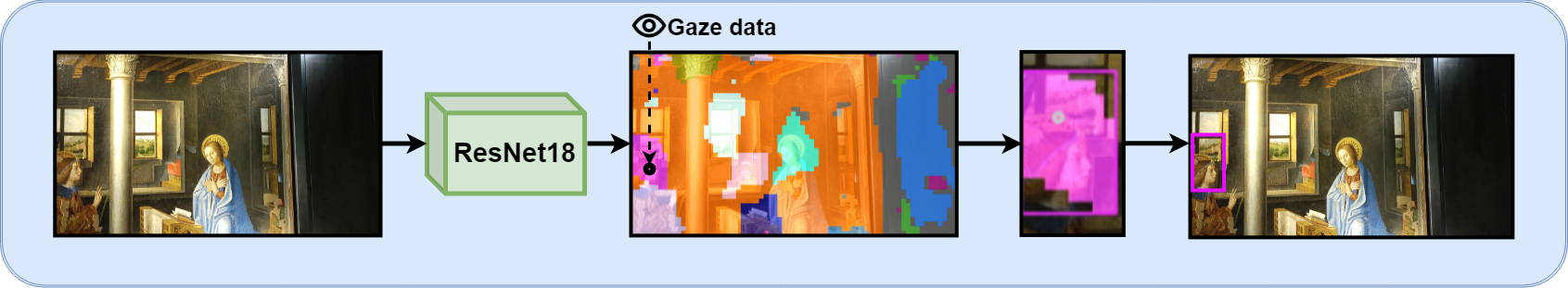}
  \caption{The proposed fully convolutional approach.}
  \label{fig:proposed_approach}
\end{figure}

The method described in the previous section has the main drawback of being slow. Indeed, processing an image at full resolution (e.g., $2272\times1278$ pixels) takes up to 168 seconds on a Tesla-K80 GPU.
To speed up the approach, inspired by~\cite{long2015fully,zhou2015learning}, we modify the trained ResNet by removing the Global Average Pooling operation and replacing it with a fully connected classifier with a $1\times1$ convolutional layer. This allows the network to predict a semantic segmentation mask of the whole image in a single step. Note that the introduced convolutional layers are initialized with the same weights of the fully connected layer in order to obtain consistent predictions. 
Given an input frame, the model outputs the class probability distributions for each pixel, which we then use to obtain a pixel-wise classification map. We up-sample the predicted mask to the original resolution to obtain a coarse segmentation mask of the whole image. We hence obtain the predicted bounding box of the attended object from the coarse segmentation map from gaze coordinates using the box fitting approach discussed in the previous section. Figure~\ref{fig:proposed_approach} presents an overview of the discussed approach.

\subsection{Finetuned Fully Convolutional Attended Object Detection} 
\label{finetuneFC}
\begin{figure}[t]
  \centering
  \includegraphics[width=0.8\linewidth]{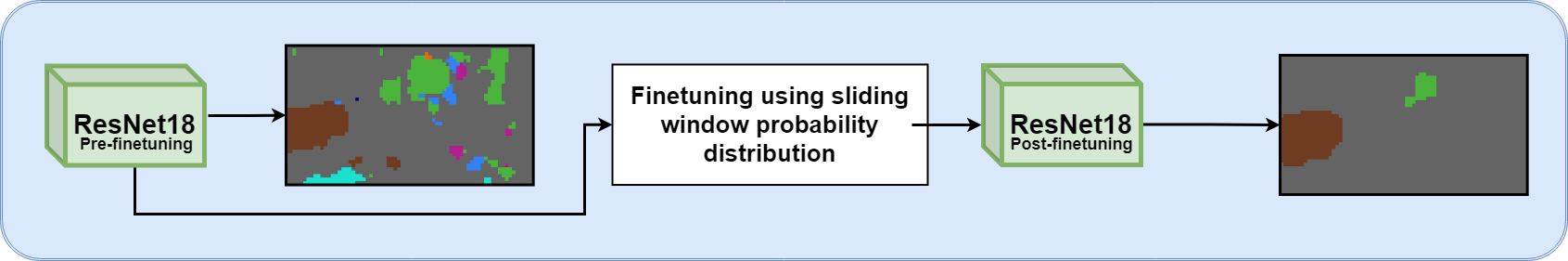}
  \caption{The proposed finetuning strategy. Note that finetuning tends to clean the masks from artifacts related to classes that are not present in the image.}
  
  \label{fig:finetuning_png}
\end{figure}

While the fully convolutional approach is much faster than the one based on the sliding window, we found the latter to be more accurate in our experiments. This is probably due to the fact that, during training, the CNN has observed only “decontextualized” image patches, rather than the full input image. We hence investigate whether fine-tuning the fully convolutional model with the output of the sliding window method can boost performance. To this aim, we use a sample of the coarse segmentation masks extracted from the training set using the sliding window approach (Section \ref{Sliding_Window_Approach}) to finetune the fully convolutional model (Figure \ref{fig:finetuning_png}). For this finetuning, we used the Kullback-Leibler Divergence score (KL) \cite{Joyce2011} to encourage the fully convolutional model to predict at each pixel location the same probability distribution predicted by the sliding window approach. 

\subsection{Fully-supervised attended object detector}   
  \label{FSupervised}
We compare the proposed weakly supervised approaches with a fully-supervised baseline based on a Faster-RCNN~\cite{ren2016faster} object detector. The detector gives as output the bounding boxes (2D coordinates) of all objects of interest present in the image. Using the 2D coordinates of the gaze, we select the bounding box which includes the gaze coordinates.

\section{ Experimental Settings and Results}
\label{experiments_section}

 \begin{table}[t]
\caption{Results obtained by the compared method. Per-column best results are reported in bold, whereas second-best results are underlined. Inference times are measured using a NVIDIA K80 GPU.} 
\centering 
\resizebox{0.94\textwidth}{!}{%
\begin{tabular}{ l c c c c } 
\hline\hline 
 \textbf{Model} & \textbf{Inference time (seconds)} & \textbf{mAP} & \textbf{mAP50}  \\
\hline 
\textbf{Sliding window approach} &168                        &\textbf{0.19}   &\textbf{0.43}                   \\
\textbf{Fully convolutional approach} &0.31        &\underline{0.18}    &0.34               \\
\textbf{Fully convolutional approach + fine-tuning} &0.31       &\textbf{0.19}   &\underline{0.41}             \\
\hline
\textbf{Faster-RCNN (baseline)~\cite{ren2016faster}}&0.80 &0.42        & 0.60              \\

\hline 
\end{tabular}}
\label{table:recap} 
\end{table}

We split the dataset into 185,457 training frames and 46,132 test frames, which is used to perform all experiments. 
The ResNet\cite{resnet} CNN has been pre-trained on ImageNet and then finetuned on training image patches sampled around gaze locations till convergence using a batch size of 8 and a learning rate of $1\mathrm{e}^{-3}$. This model was then used to evaluate both the sliding windows approach (Section~\ref{Sliding_Window_Approach}) and the fully convolutional approach (Section~\ref{conv}). To finetune the fully convolutional approach, we selected 1659 frames from the training set where the attended object was different than the ``other" class.
The finetuning using the KL divergence has been performed till convergence with a batch size of 1 and a learning rate of $1\mathrm{e}^{-3}$.

\begin{figure}[t]
  \centering
  \includegraphics[width=0.8\linewidth]{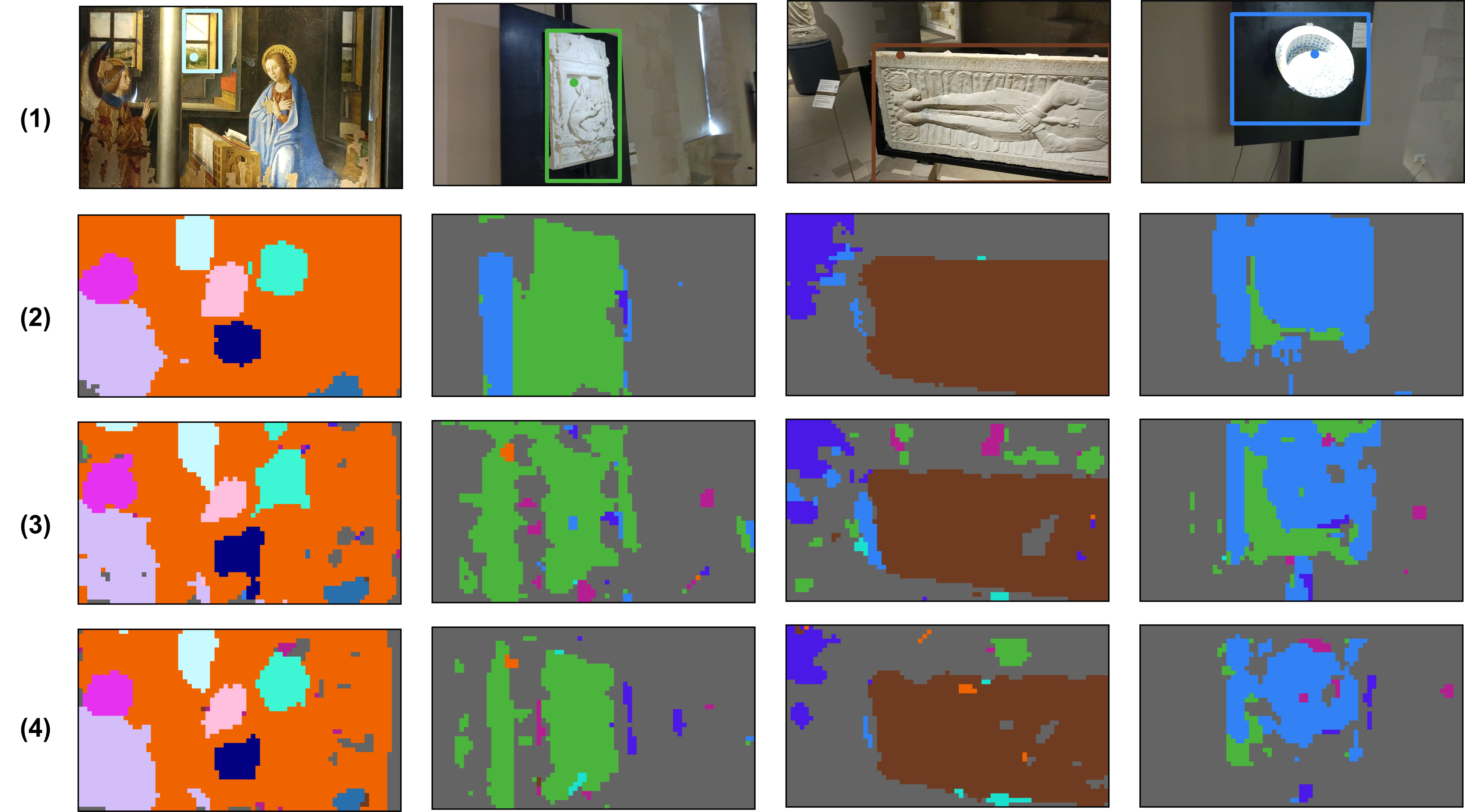}
  \caption{Qualitative examples of the compared approaches. (1) Input image with the object detections obtained by the fully convolutional +finetuning approach. (2) Output mask of the sliding window approach. (3) Mask produced by the fully convolutional approach without finetuning. (4) Mask produced by the fully convolutional method after finetuning.}
  \label{fig:final_recap}
\end{figure}

The performances obtained with the four compared methods are reported in Table \ref{table:recap}. 
We evaluated the methods using the mAP and mAP50 measures~\cite{lin2015microsoft}, which are commonly used for the evaluation of object detectors. We also computed inference times using a NVIDIA K80 GPU. The sliding window approach allows to achieve a decent performance with an mAP of 0.19 and an mAP50 of 0.43, relying only on gaze and attended object frame-wise annotations as supervision. Despite these results, the approach cannot be exploited in practice due to the high inference time of 168 seconds per image. The proposed fully convolutional approach allows to greatly reduce inference time to 0.31 seconds per image, however it achieves a weaker performance with an mAP 0.18 and an mAP50 of 0.34. Fine-tuning the fully convolutional model with the proposed optimization procedure allows to achieve a  performance similar to one obtained with the sliding window approach, with an mAP of 0.19 and an mAP50 of 0.41, while retaining the reduced inference time of 0.31 seconds per image. 
The fully supervised object detector obtains better performance than the weakly supervised method (0.42 mAP vs 0.19 mAP of the finetuned fully convolutional approach), but it requires labeled bounding boxes at training time, which would greatly affect the deployment cost of the final system. It is worth noting that, besides requiring less supervision, the finetuned fully convolutional approach also brings significant time savings, with a runtime reduced of about $62\%$ with respect to the fully superivised model (0.31s vs 0.8s). Qualitative results are shown in Figure~\ref{fig:final_recap}. The examples show as the masks obtained by the sliding window approach (Figure \ref{fig:final_recap} row (2)) are the most homogeneous and compact, while those of the fully convolutional approach (Figure~\ref{fig:final_recap} row (3)) are more noisy. After finetuning (Figure~\ref{fig:final_recap} row (4)), it is possible to observe an improvement due to the removal of spurious pixels.

\section{Conclusion}
\label{conclusion_section}
We investigated the problem of weakly supervised attended object detection in cultural sites starting from the consideration that manually labeling images with bounding boxes around all objects is a time-consuming process. To study the problem, we acquired a new dataset of egocentric videos in a cultural site including 15 objects of interest and a negative class (``other"). The dataset is composed of egocentric videos and gaze signals, and includes 178,977 frames manually labelled with bounding box annotations for evaluation purposes. We investigated three different approaches to detect attended objects in a weakly supervised way, by leaning only on gaze data and a frame-level label indicating the class of the attended object. Results show that our fully convolutional approach, after a fine-tuning stage, obtains good performance and has a lower inference time (0.31s) with respect to a fully supervised detector. The proposed dataset can be also used to address the task of fully supervised object detection in cultural sites. Future works will focus on improving our approach to obtain comparable performance in terms of accuracy with respect to a fully-supervised method.

\section*{Acknowledgements}

This research has been supported by Next Vision\footnote{Next Vision: \href{https://www.nextvisionlab.it/}{https://www.nextvisionlab.it/}} s.r.l., by the project VALUE (N. 08CT6209090207 - CUP G69J18001060007) - PO FESR 2014/2020 - Azione 1.1.5., and by Research Program  Pia.ce.ri. 2020/2022 Linea 2 - University of Catania.

\section*{\uppercase{Supplementary material}}
This document is intended for the convenience of the reader and reports additional information about the collection of the proposed dataset, as well as implementation details of the performed experiments.  The remainder of this document is organized as follows. Section~\ref{Mixed} reports additional details about data collection. Section~\ref{experiment} provides implementation details of the compared methods and additional qualitative results.

\section{Mixed reality application for data collection}
\label{Mixed}
\begin{figure}[h!]
  \centering
  \includegraphics[width=1\linewidth]{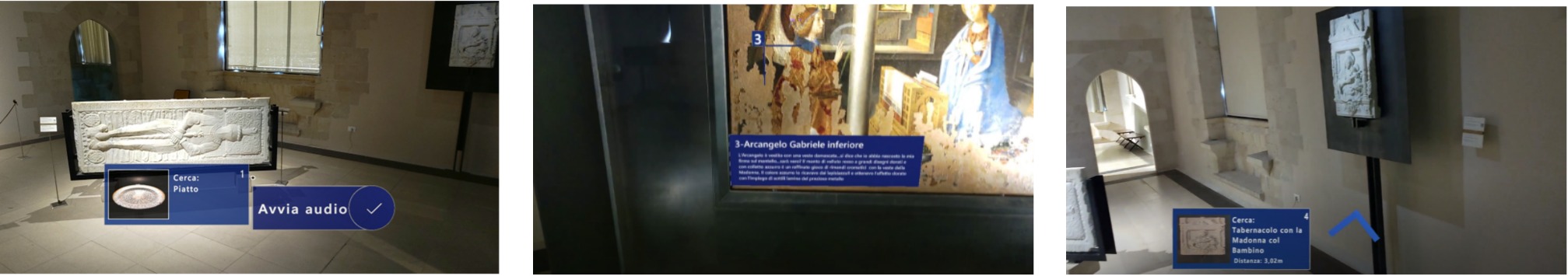}
  \caption{Example of the developed application used for the data acquisition. In blue, the elements shown in mixed reality.}
  \label{fig:app}
\end{figure}

In order to acquire the dataset, we developed a Hololens 2 application which allows two acquisition modalities: “tour” and “free”. In the “tour" mode, the application proposes to the visitors a randomly generated path to follow to se artworks among all of the 15 considered objects of interest. The visitors have to follow the suggested path, stop in front of each object and listen to the audio description. in the “free" mode the user is free to move in the space and look at the different object of interest without any suggestions. Before starting the tour, it is possible to start the video stream capture through the voice command “Record". This command starts the video streaming and initializes the log files.\\\\
The application has been developed using Unity and MRTK~\cite{MRTK}, an open source toolkit released by Microsoft community. Figure~\ref{fig:app} shows some example of the developed application.

\section{Experiment and Implementation details}
\label{experiment}
The dataset was divided into training and test sets as shown in Table~\ref{table:SUPPLEMNETARY}. All the performed experiments use this partition.

\begin{table}[h!]
\caption{Info about the videos of the dataset } 
\centering
\label{table:SUPPLEMNETARY}
\resizebox{0.8\textwidth}{!}{%
\begin{tabular}{c c c c c c}
\hline\hline 
\textbf{ID}  & \textbf{SUBJECT} & \textbf{TYPE} & \textbf{VIDEO NAME} & \textbf{SPLIT} & \textbf{MINUTES} \\ 
\hline
1    & 1 & Tour & 2021-03-02\_01-57-51\_031\_Video & Train &12.8\\ 
3    & 4 & Tour & 2021-03-02\_02-56-17\_227\_Video & Train &12.1\\ 
4     & 2 & Tour & 2021-07-12\_02-32-22\_575\_Video  & Train &11.2\\ 
5     & 3 & Tour & 2021-07-12\_02-56-41\_326\_Video  & Train &11.4\\ 
7     & 6 & Tour & 2021-07-09\_11-16-49\_610\_Video  & Train &12.3\\ 
8     & 7 & Tour & 2021-07-09\_11-38-43\_958\_Video  & Train &11.7\\ 
9     & 2 & Free & 2021-07-12\_08-40-16\_306\_Video  & Train &2.3\\ 
10    & 2 & Free & 2021-07-12\_01-33-09\_637\_Video  & Train &7.1\\ 
11    & 3 & Free & 2021-07-12\_01-45-18\_300\_Video  & Train &6.4\\ 
12    & 5 & Free & 2021-07-12\_01-56-35\_931\_Video  & Train &8.5\\ 
14    & 7 & Free & 2021-07-12\_02-19-09\_623\_Video  & Train &7.1\\ 
\hline
2    & 2 & Tour & 2021-03-02\_02-39-14\_296\_Video & Test  & 8.1\\ 
6     & 5 & Tour & 2021-07-12\_03-21-17\_102\_Video  & Test & 13.1 \\ 
13    & 6 & Free & 2021-07-12\_02-09-55\_623\_Video  & Test  &4,4\\ 
\hline
\end{tabular}%
}
\end{table}

\begin{figure}[t]
  \centering
  \includegraphics[width=0.8\linewidth]{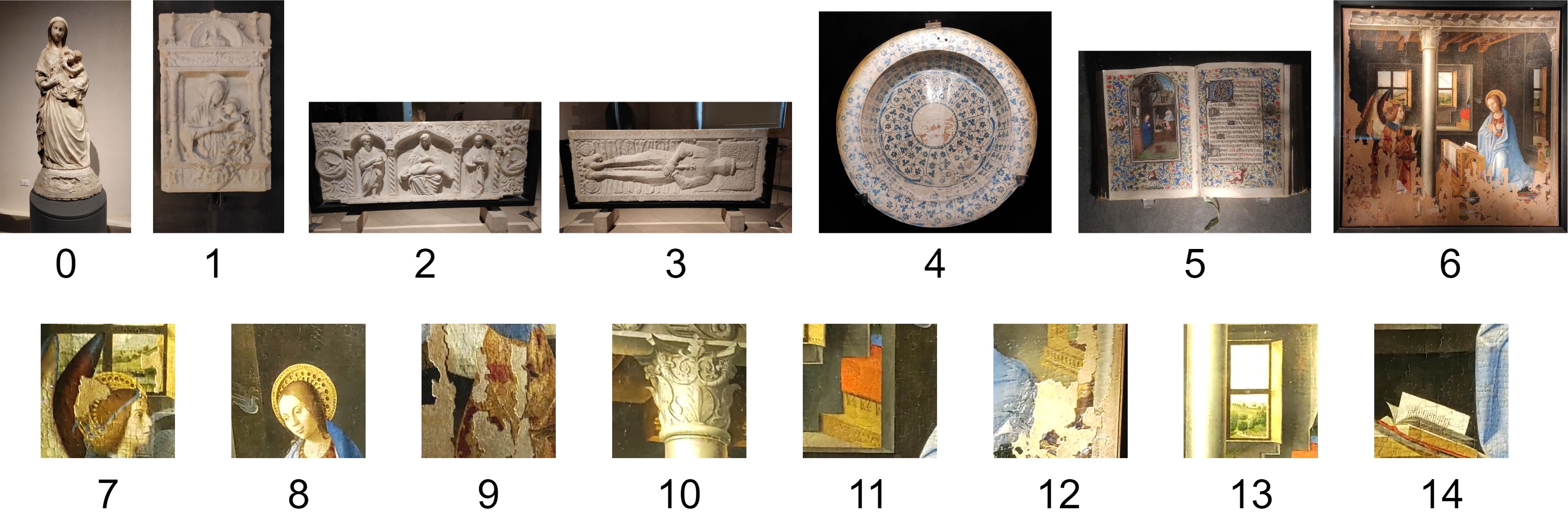}
  \caption{The considered objects of interest belonging to the dataset. From 0 to 14:  ``Madonna del Cardillo",
   ``Tabernacolo con la Madonna col Bambino",
  ``Lastra tombale di Giovanni Cabastida retro", ``Lastra tombale di Giovanni Cabastida fronte",
  ``Piatto fondo",``Libro d'Ore miniato",  ``Annunciazione",
  ``Dettaglio Arcangelo Gabriele superiore",
  ``Dettaglio Vergine"
  ``Arcangelo Gabriele parte inferiore",
  ``Capitello",
  ``Letto",
  ``Devoto in basso a dx",
  ``Finestra centrale",
  ``Sacre Scritture". The ``Other" class is related to all the other elements of the scene. }
  
  \label{fig:dataset_png}
\end{figure}

\subsection{Fully-supervised attended object detector}
    
\begin{figure}[h!]
  \centering
  \includegraphics[width=1\linewidth]{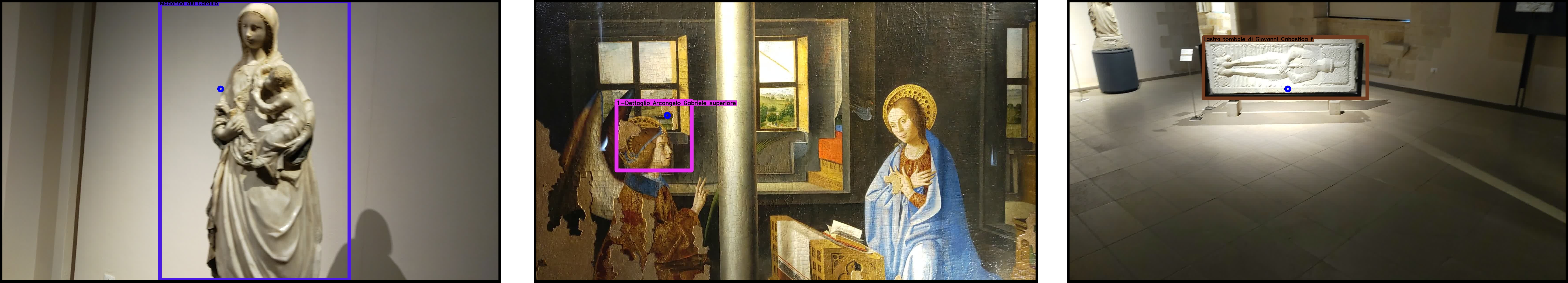}
  \caption{Examples of attended object detection. The blue circle indicates the gaze, whereas the colored box indicates the attended object detected.}
  \label{fig:detection}
\end{figure}

In order to investigate the problem of attended object detection, as baseline, we trained a object detector (Faster R-CNN~\cite{ren2016faster}). In particular, we trained the model using 17,8977 images, with 713,097 bounding box annotations. We also added frames and annotations from the dataset at the following url: \url{https://iplab.dmi.unict.it/EGO-CH/}. 
During the inference phase, we considered the gaze signals to select the attended object. The obtained results are shown in Table~\ref{table:apattendeddedetcion}. Figure \ref{fig:dataset_png} show an example of the different classes
Results show that identifying the details contained in the object of interest “Annunciazione" is very challenging (classes 7-14), both because of their small size and because they are contained within a larger picture to be recognised as whole. Figure~\ref{fig:detection} shows qualitative results.

\begin{table}[h!]
\caption{mAP of  Attended Object of interest detection} 
\centering 
\resizebox{0.9\textwidth}{!}{%
\begin{tabular}{lllllllllllllllll}
\hline\hline 
\textbf{Class}& 0    & 1    & 2    & 3    & 4    & 5    & 6    & 7     & 8     & 9    & 10    & 11   & 12   & 13   & 14  &\textbf{All}   \\
\hline 
\textbf{AP}    & 0.72 & 0.75 & 0.83 & 0.79 & 0.74 & 0.87 & 0.74 & 0.01  & 0.08  & 0.18 & 0.02  & 0.10 & 0.10 & 0.22 & 0.13 &0.42 \\
\textbf{AP50}  & 0.90 & 0.90 & 0.98 & 0.87 & 0.90 & 0.90 & 0.89 & 0.06  & 0.33  & 0.41 & 0.05  & 0.27 & 0.31 & 0.64 & 0.64 &0.60 \\

\hline
\end{tabular}}

\label{table:apattendeddedetcion} 
\end{table}

\subsection{Attended object with Sliding windows approach}

\begin{figure}[t]
  \centering
  \includegraphics[width=0.8\linewidth]{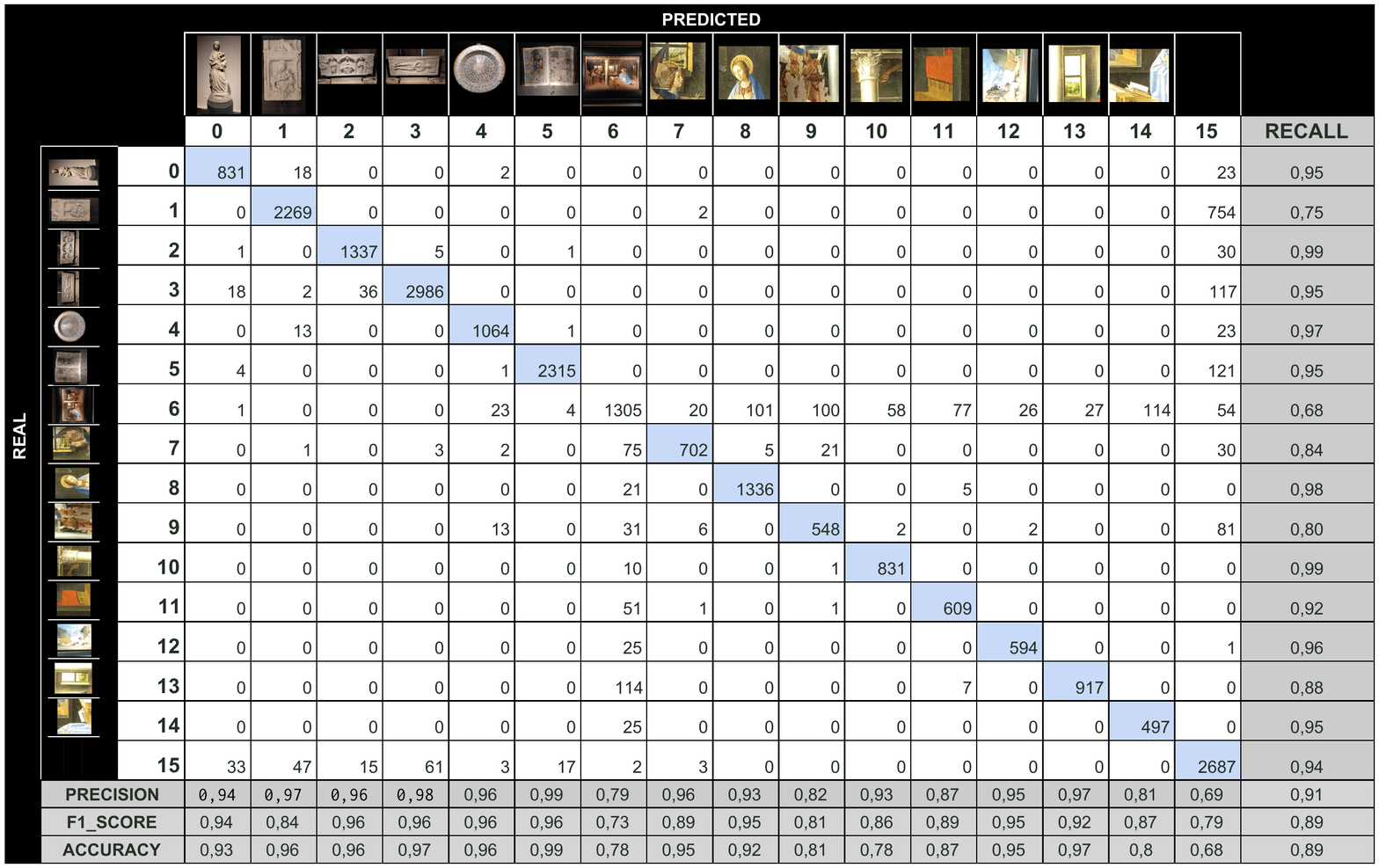}
  \caption{Classification confusion matrix.}
  \label{fig:classification}
\end{figure}

\begin{figure}[h!]
  \centering
  \begin{subfigure}[b]{0.38\linewidth}
    \includegraphics[width=\linewidth]{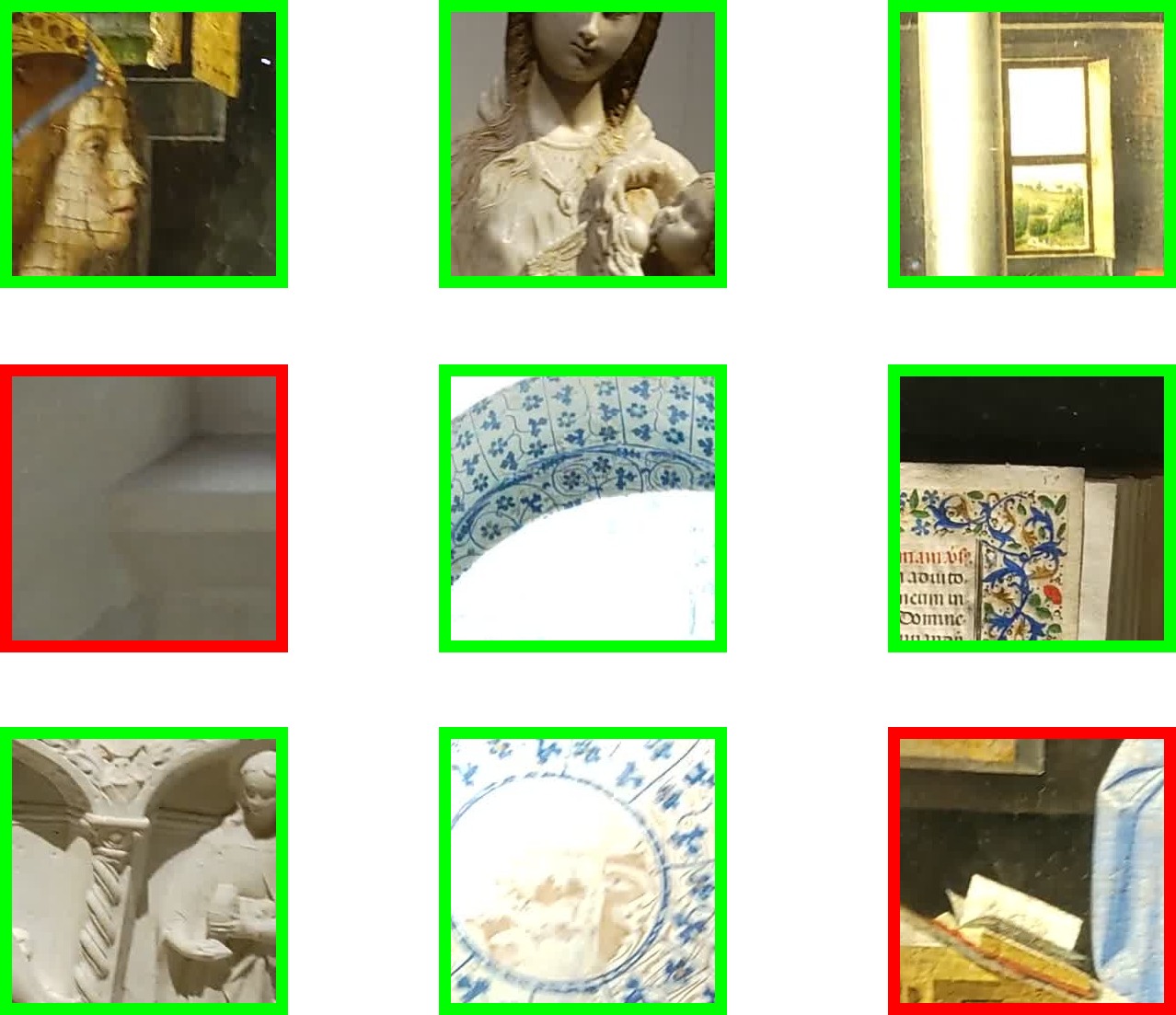}
    \caption{}
  \end{subfigure}
  \begin{subfigure}[b]{0.48\linewidth}
    \includegraphics[width=\linewidth]{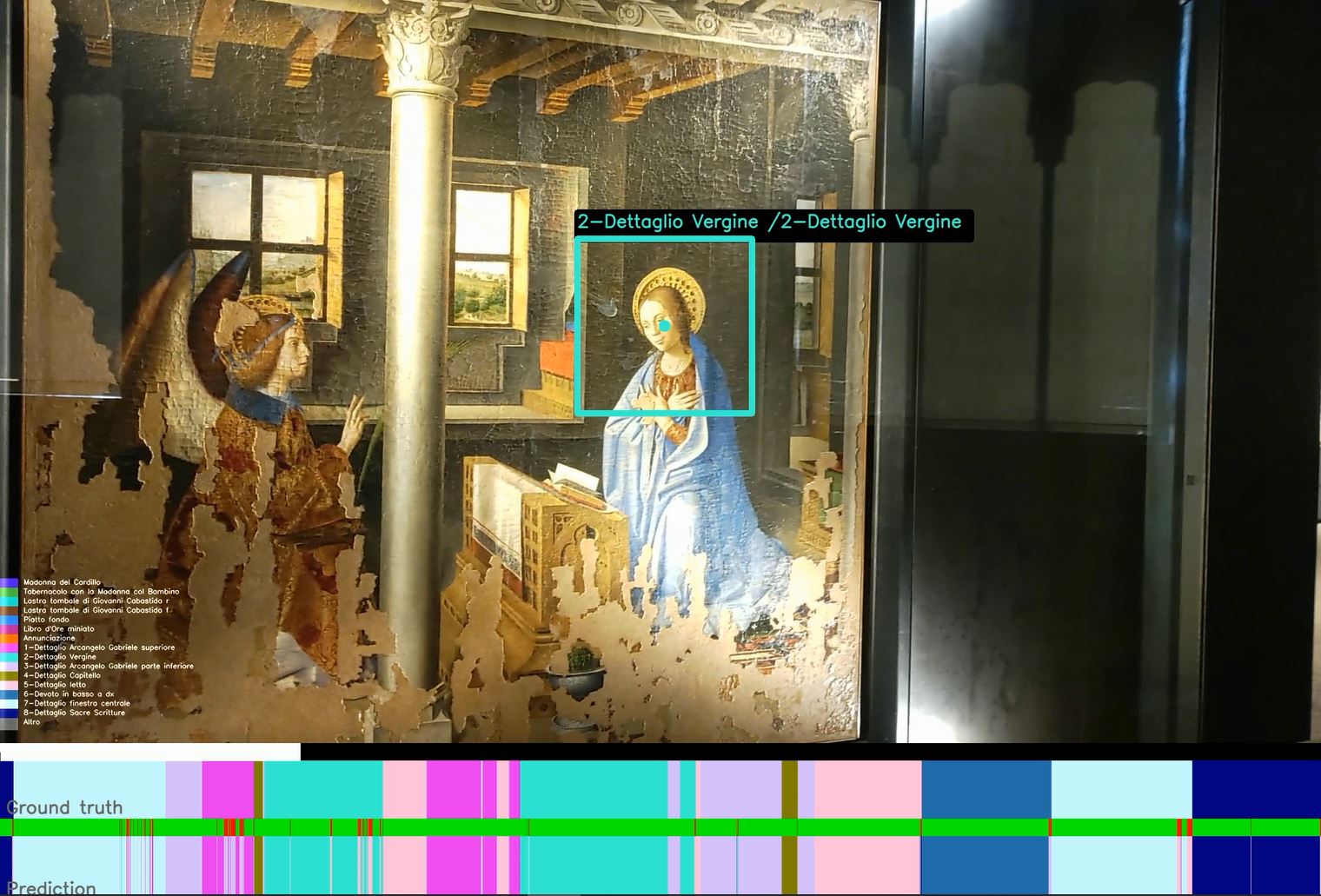}
    \caption{}
  \end{subfigure}
  \caption{Attended object classification. (a) Gaze patch classification. Correct predictions are reported in green whereas wrong predictions are reported in red. (b) The box represents the patch centered on the gaze (colored point). The color indicates the object class. }
  \label{fig:patch}
\end{figure}

\begin{figure}[h!]
  \centering
  \includegraphics[width=0.8\linewidth]{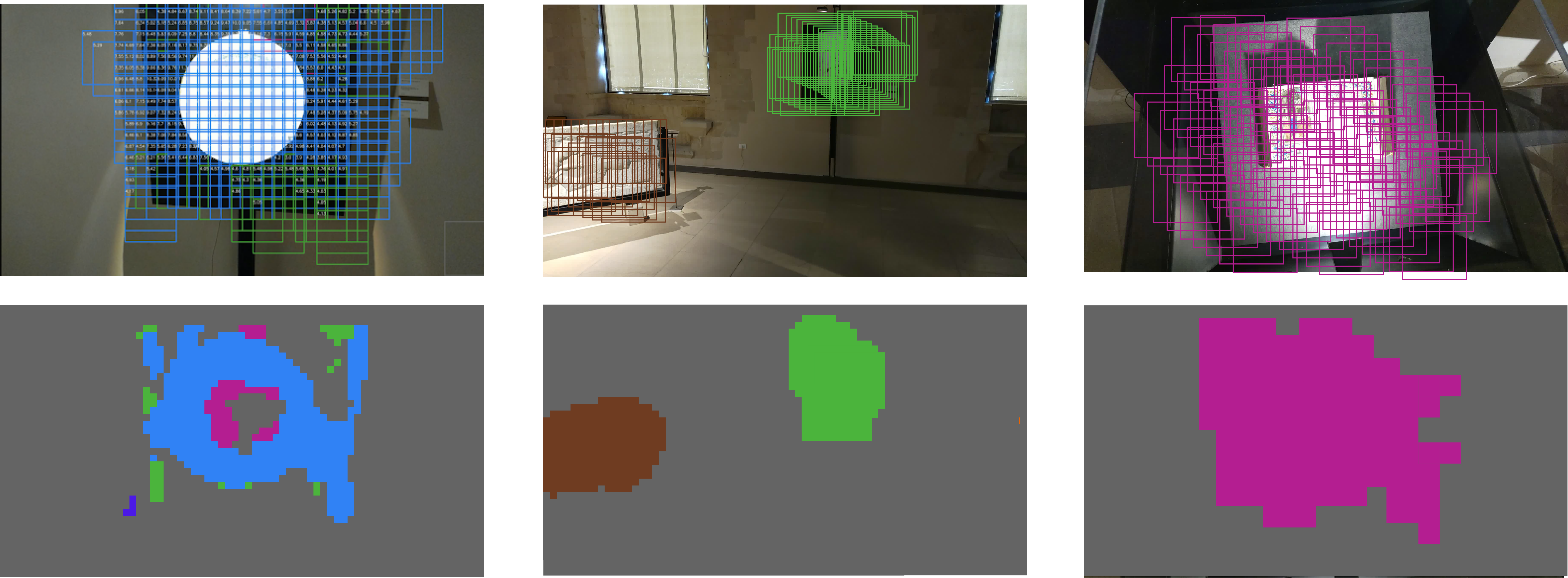}
  \caption{Example of sliding window outputs.}
  \label{fig:slidinr_handmade}
\end{figure}

For the classification step of the Sliding Window approach we trained an image classifier (ResNet18~\cite{resnet}). As supervision, we used only the punctual annotation relating to the gaze. We discarded the frames in which the reprojection of the gaze falls outside the bounding box of the frame, obtaining 17,5985 training frames and 22,719 test frames. The model was trained using patches of $300\times300$ pixels extracted around the gaze. Figure~\ref{fig:classification} reports the confusion matrix as well as the accuracy, recal and precision of the model. Figure~\ref{fig:patch} shows some qualitative results. At test time, a $300\times300$ pixels window slides on the input frame, giving as output a tensor of size: $40 \times 71 \times 1$. This method lacks in time-speed. Indeed, to process a single frame 168 seconds are needed (using a NVIDIA Tesla K80). The slowness of this approach is due to the fact that two adjacent patches share about $90\%$ of the pixels. Some qualitative results are shown in Figure~\ref{fig:slidinr_handmade}.  Achieved result are shown in the Table~\ref{table:recap_suppl}.

\subsection{Fully Convolutional attended object detection (Pre finetuning)}

Figure~\ref{fig:prevspost} (a) shows some example of crowed prediction of the model pre-finetuning.
The masks produced by this approach highlight how the network fails in presence of small objects, with many objects in the scene or when there are shaky frames. The aforementioned figure highlights that there are different cases where multiple pixels are classified as belonging to one class instead of the negative class “other". Achieved result are shown in the Table~\ref{table:recap_suppl}.

\subsection{Convolutional attended object detection (Post finetuning)}
\begin{figure}[h!]
  \centering
  \begin{subfigure}[b]{1\linewidth}
    \includegraphics[width=\linewidth]{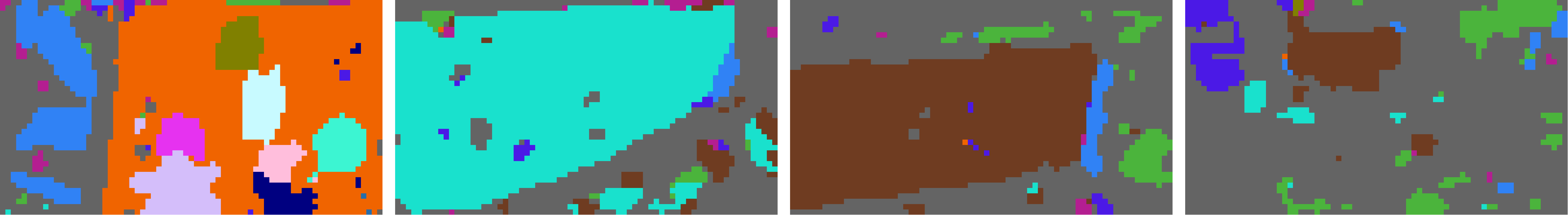}
    \caption{Predictions pre-finetuning.}
    \
  \end{subfigure}
  \begin{subfigure}[b]{1\linewidth}
    \includegraphics[width=\linewidth]{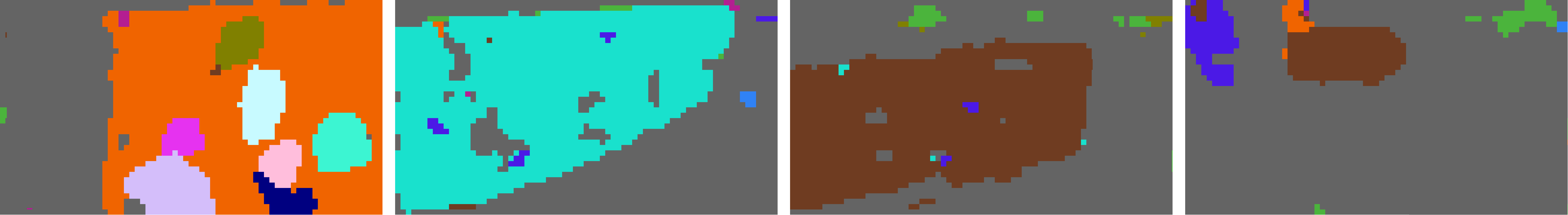}
    \caption{Predictions post-finetuning.}
  \end{subfigure}
  \caption{Example of segmentation mask pre and post finetuning.}
  \label{fig:prevspost}
\end{figure}

Qualitative results are shown in Figure~\ref{fig:prevspost} (b). Achieved result are shown in Table~\ref{table:recap_suppl}.

 \begin{table}[t]
\caption{Results obtained by the compared method. Per-column best results are reported in bold, whereas second-best results are underlined. Inference times are measured using a NVIDIA K80 GPU.} 
\centering 
\resizebox{1.0\textwidth}{!}{%
\begin{tabular}{ l c c c c } 
\hline\hline 
 \textbf{Model} & \textbf{Inference time (seconds)} & \textbf{mAP} & \textbf{mAP50}  \\
\hline 
\textbf{Sliding window approach} &168                        &\textbf{0.19}   &\textbf{0.43}                   \\
\textbf{Fully convolutional approach} &0.31        &\underline{0.18}    &0.34               \\
\textbf{Fully convolutional approach + fine-tuning} &0.31       &\textbf{0.19}   &\underline{0.41}             \\
\hline
\textbf{Faster-RCNN (baseline)~\cite{ren2016faster}}&0.80 &0.42        & 0.60              \\

\hline 
\end{tabular}}
\label{table:recap_suppl} 
\end{table}

%
%
%
%

\newpage
\newpage
\clearpage
 \bibliographystyle{splncs04}
 \bibliography{mybibliography.bib}

\begin{thebibliography}{10}
\providecommand{\url}[1]{\texttt{#1}}
\providecommand{\urlprefix}{URL }
\providecommand{\doi}[1]{https://doi.org/#1}

\bibitem{MRTK}
{MRTK}.
  \url{https://docs.microsoft.com/it-it/windows/mixed-reality/mrtk-unity/?view=mrtkunity-2021-05/}

\bibitem{Bearman2016WhatsTP}
Bearman, A.L., Russakovsky, O., Ferrari, V., Fei-Fei, L.: What's the point:
  Semantic segmentation with point supervision. In: European Conference on
  Computer Vision. pp. 549--565 (2016)

\bibitem{chen2016semantic}
Chen, L.C., Papandreou, G., Kokkinos, I., Murphy, K., Yuille, A.L.: Deeplab:
  Semantic image segmentation with deep convolutional nets, atrous convolution,
  and fully connected {CRFS}. Transactions on Pattern Analysis and Machine
  Intelligence  \textbf{40}(4),  834--848 (2018)

\bibitem{cheng2021pointlysupervised}
Cheng, B., Parkhi, O., Kirillov, A.: Pointly-supervised instance segmentation.
  In: ArXiv (2021)

\bibitem{NoisyArt}
Chiaro, R.D., Bagdanov, A.D., Bimbo, A.: Noisyart: A dataset for
  webly-supervised artwork recognition. In: International Conference on
  Computer Vision Theory and Applications (2019)

\bibitem{6338939}
Farabet, C., Couprie, C., Najman, L., LeCun, Y.: Learning hierarchical features
  for scene labeling. Transactions on Pattern Analysis and Machine Intelligence
   \textbf{35}(8),  1915--1929 (2013)

\bibitem{redmon2018yolov3}
Farhadi, A., Redmon, J.: Yolov3: An incremental improvement. In: Computer
  Vision and Pattern Recognition. vol.~1804 (2018)

\bibitem{Furnari20214021}
Furnari, A., Farinella, G.: {Rolling-Unrolling LSTMs} for action anticipation
  from first-person video. Transactions on Pattern Analysis and Machine
  Intelligence  \textbf{43}(11),  4021--4036 (2021)

\bibitem{Furnari2016474}
Furnari, A., Farinella, G., Battiato, S.: Temporal segmentation of egocentric
  videos to highlight personal locations of interest. In: Lecture Notes in
  Computer Science. vol. 9913 LNCS, pp. 474--489 (2016)

\bibitem{garcia2020dataset}
Garcia, N., Ye, C., Liu, Z., Hu, Q., Otani, M., Chu, C., Nakashima, Y.,
  Mitamura, T.: A dataset and baselines for visual question answering on art.
  In: European Conference on Computer Vision Workshops. pp. 92--108 (2020)

\bibitem{fastrcnn}
Girshick, R.B.: Fast {{R-CNN}}. In: International Conference on Computer
  Vision. pp. 1440--1448 (2015)

\bibitem{rcnn}
Girshick, R.B., Donahue, J., Darrell, T., Malik, J.: Rich feature hierarchies
  for accurate object detection and semantic segmentation. Computer Vision and
  Pattern Recognition pp. 580--587 (2014)

\bibitem{hariharan2014simultaneous}
Hariharan, B., Arbel{\'a}ez, P., Girshick, R., Malik, J.: Simultaneous
  detection and segmentation. In: European Conference on Computer Vision. pp.
  297--312 (2014)

\bibitem{resnet}
He, K., Zhang, X., Ren, S., Sun, J.: Deep residual learning for image
  recognition. In: Computer Vision and Pattern Recognition. pp. 770--778 (2016)

\bibitem{jimaging8020018}
Ibrahim, B.I.E., Eyharabide, V., Le~Page, V., Billiet, F.: Few-shot object
  detection: Application to medieval musicological studies. Journal of Imaging
  \textbf{8}(2) (2022)

\bibitem{Joyce2011}
Joyce, J.M.: Kullback-Leibler Divergence, pp. 720--722. Berlin, Heidelberg
  (2011)

\bibitem{6751187}
Karthikeyan, S., Jagadeesh, V., Shenoy, R., Ecksteinz, M., Manjunath, B.: From
  where and how to what we see. In: International Conference on Computer
  Vision. pp. 625--632 (2013)

\bibitem{7298944}
Karthikeyan, S., Ngo, T., Eckstein, M., Manjunath, B.: Eye tracking assisted
  extraction of attentionally important objects from videos. In: Computer
  Vision and Pattern Recognition. pp. 3241--3250 (2015)

\bibitem{koniusz2018museum}
Koniusz, P., Tas, Y., Zhang, H., Harandi, M., Porikli, F., Zhang, R.: Museum
  exhibit identification challenge for the supervised domain adaptation and
  beyond. In: European Conference on Computer Vision (September 2018)

\bibitem{6909437}
Li, Y., Hou, X., Koch, C., Rehg, J.M., Yuille, A.L.: The secrets of salient
  object segmentation. In: Computer Vision and Pattern Recognition). pp.
  280--287 (2014)

\bibitem{retinanet}
Lin, T.Y., Goyal, P., Girshick, R.B., He, K., Doll{\'a}r, P.: Focal loss for
  dense object detection. Transactions on Pattern Analysis and Machine
  Intelligence  (2020)

\bibitem{lin2015microsoft}
Lin, T.Y., Maire, M., Belongie, S.J., Hays, J., Perona, P., Ramanan, D.,
  Doll{\'a}r, P., Zitnick, C.L.: Microsoft {{COCO}}: Common objects in context.
  In: European Conference on Computer Vision (2014)

\bibitem{liu2016ssd}
Liu, W., Anguelov, D., Erhan, D., Szegedy, C., Reed, S., Fu, C.Y., Berg, A.C.:
  Single shot multibox detector. In: European Conference on Computer Vision
  (2016)

\bibitem{5459254}
Mishra, A., Aloimonos, Y., Fah, C.L.: Active segmentation with fixation. In:
  International Conference on Computer Vision. pp. 468--475 (2009)

\bibitem{Papadopoulos2014TrainingOC}
Papadopoulos, D.P., Clarke, A.D.F., Keller, F., Ferrari, V.: Training object
  class detectors from eye tracking data. In: European Conference on Computer
  Vision (2014)

\bibitem{PATHAK202174}
Pathak, R., Saini, A., Wadhwa, A., Sharma, H., Sangwan, D.: An object detection
  approach for detecting damages in heritage sites using 3-d point clouds and
  2-d visual data. Journal of Cultural Heritage  \textbf{48},  74--82 (2021)

\bibitem{ragusa2019egocentricpoint}
Ragusa, F., Furnari, A., Battiato, S., Signorello, G., Farinella, G.M.:
  Egocentric point of interest recognition in cultural sites. In: International
  Conference on Computer Vision Theory and Applications (VISAPP) (2019)

\bibitem{EGO-CH}
Ragusa, F., Furnari, A., Battiato, S., Signorello, G., Farinella, G.M.:
  {EGO-CH:} dataset and fundamental tasks for visitors behavioral understanding
  using egocentric vision. Pattern Recognition Letters  \textbf{131},  150--157
  (2020)

\bibitem{redmon2016look}
Redmon, J., Divvala, S.K., Girshick, R.B., Farhadi, A.: You only look once:
  Unified, real-time object detection. In: Computer Vision and Pattern
  Recognition. pp. 779--788 (2016)

\bibitem{ren2016faster}
Ren, S., He, K., Girshick, R., Sun, J.: Faster {R-CNN}: Towards real-time
  object detection with region proposal networks. In: Advances in Neural
  Information Processing Systems (2015)

\bibitem{10.1145/3092832}
Seidenari, L., Baecchi, C., Uricchio, T., Ferracani, A., Bertini, M., Bimbo,
  A.D.: Deep artwork detection and retrieval for automatic context-aware audio
  guides. Transactions on Multimedia Computing, Communications, and
  Applications  \textbf{13} (2017)

\bibitem{long2015fully}
Shelhamer, E., Long, J., Darrell, T.: Fully convolutional networks for semantic
  segmentation. Transactions on Pattern Analysis and Machine Intelligence
  \textbf{39},  640--651 (2017)

\bibitem{See_Better}
Subramanian, R., Yanulevskaya, V., Sebe, N.: Can computers learn from humans to
  see better? inferring scene semantics from viewers' eye movements. In:
  International Conference on Multimedia (ACM). p. 33–42 (2011)

\bibitem{2021}
Wang, Y., Hou, J., Hou, X., Chau, L.P.: A self-training approach for
  point-supervised object detection and counting in crowds. Transactions on
  Image Processing  \textbf{30},  2876–2887 (2021)

\bibitem{yoo2019pseudoedgenet}
Yoo, I., Yoo, D., Paeng, K.: Pseudoedgenet: Nuclei segmentation only with point
  annotations. In: Medical Image Computing and Computer Assisted Intervention
  (MICCAI). pp. 731--739 (2019)

\bibitem{DBLP:conf/cvpr/YunPSZB13}
Yun, K., Peng, Y., Samaras, D., Zelinsky, G.J., Berg, T.L.: Studying
  relationships between human gaze, description, and computer vision. In:
  Computer Vision and Pattern Recognition. pp. 739--746 (2013)

\bibitem{zhou2015learning}
Zhou, B., Khosla, A., Lapedriza, {\`A}., Oliva, A., Torralba, A.: Learning deep
  features for discriminative localization. In: Computer Vision and Pattern
  Recognition. pp. 2921--2929 (2016)

\end{thebibliography}
\end{document}